\documentclass[letterpaper, 10 pt]{ieeeconf}  

\usepackage{times}
\usepackage{epsfig}
\usepackage{graphicx}
\usepackage{amsmath}
\usepackage{amssymb}
\usepackage{flexisym}
\usepackage{booktabs}
\usepackage{multirow}
\usepackage{array}
\usepackage{tabularx}
\usepackage{ragged2e}
\usepackage{multicol}
\usepackage{lineno}
\usepackage{fixltx2e}
\usepackage{trimclip}
\usepackage{scalerel}
\usepackage{subfigure}
\usepackage{gensymb}
\usepackage{subfigure}

\IEEEoverridecommandlockouts                              

\title{\LARGE \bf
A Consistency-Based Loss for Deep Odometry Through Uncertainty Propagation
}

\author{Hamed Damirchi$^*$, Rooholla Khorrambakht$^*$, Hamid D. Taghirad$^*$, and Behzad Moshiri$^{**}$ 
\thanks{$^*$Advanced Robotics and Automated Systems (ARAS),
Faculty of Electrical Engineering, K. N. Toosi University of
Technology, Tehran, Iran.}
\thanks{$^{**}$School of ECE, University College of Engineering, University of Tehran, Tehran, Iran}
\thanks{E-mails: {\tt\small hdamirchi@email, r.khorrambakht@email, taghirad@kntu.ac.ir., moshiri@ut.ac.ir}}%
}

\begin{document}

\maketitle
\thispagestyle{empty}
\pagestyle{empty}

\begin{abstract}

The incremental poses computed through odometry can be integrated over time to calculate the pose of a device with respect to an initial location. The resulting global pose may be used to formulate a second, consistency based, loss term in a deep odometry setting. In such cases where multiple losses are imposed on a network, the uncertainty over each output can be derived to weigh the different loss terms in a maximum likelihood setting. However, when imposing a constraint on the integrated transformation, due to how only odometry is estimated at each iteration of the algorithm, there is no information about the uncertainty associated with the global pose to weigh the global loss term. In this paper, we associate uncertainties with the output poses of a deep odometry network and propagate the uncertainties through each iteration. Our goal is to use the estimated covariance matrix at each incremental step to weigh the loss at the corresponding step while weighting the global loss term using the compounded uncertainty. This formulation provides an adaptive method to weigh the incremental and integrated loss terms against each other, noting the increase in uncertainty as new estimates arrive. We provide quantitative and qualitative analysis of pose estimates and show that our method surpasses the accuracy of the state-of-the-art Visual Odometry approaches. Then, uncertainty estimates are evaluated and comparisons against fixed baselines are provided. Finally, the uncertainty values are used in a realistic example to show the effectiveness of uncertainty quantification for localization.

\end{abstract}

\section{INTRODUCTION}

Odometry refers to the incremental localization of a device using sensors such as cameras, IMUs, radars, etc. This method of localization has been used in both single--modal \cite{wang2017deepvo} and multi--modal \cite{chen2019selectivevio} settings in various fields such as robotics \cite{yousif2015overview}, self-driving vehicles \cite{howard2008real} and planetary exploration rovers \cite{cheng2005visual}. Over the last decade, due to the increase in utilization of such pipelines in everyday applications, the necessity of uncertainty communication has increased for safety and reliability reasons \cite{abdar2020review}. The benefits of uncertainty quantification are not limited to uncertainty communication. In classical pose-graph based localization methods, the odometry estimates are used as constraints in between nodes of a Bayesian network where each node represents the location of the device. Although each edge is commonly given a constant covariance matrix or uses photometric errors as a heuristic for uncertainty, it has been shown \cite{wang2018end}, that estimating an uncertainty for each of the edges allows for a considerable improvement over the accuracy of the pose estimation pipeline.

Deep learning has shown to be an adequate method of learning representations from which uncertainty about a particular output can be estimated \cite{abdar2020review}. Kendal, et. al. \cite{kendall2017uncertainties}, categorized the total uncertainty of a network about an output into aleatoric and epistemic uncertainties where the aleatory variability of the output corresponds to the heteroscedastic noise in the data. The epistemic uncertainty is the result of imperfect training data (e.g. insufficient training samples) and describes the confidence of the model about it's knowledge of a certain data point. Therefore, epistemic uncertainty can be reduced by providing the model with more task representative data, whereas uncertainties are categorized as aleatory if the model cannot reduce them using more training data. 
Pragmatically, Gal, et al. \cite{gal2016dropout} used dropout variational inference to calculate the epistemic uncertainty about the output of the network and Kendal~\cite{kendall2017uncertainties} derives the aleatoric uncertainty about a datapoint through the network itself and proposes to incorporate the estimated covariance matrix within a maximum likelihood setting. Finally, the total uncertainty is calculated by summing the aleatory and epistemic uncertainties together.

Although estimating the uncertainty about the pose output from an odometry network has been formulated both in end-to-end and hybrid systems, no long-term constraints are imposed on the networks trained to deliver the uncertainty estimates. The current literature either imposes consistency constraints without the inclusion of the uncertainty \cite{saputra2019learning} or focuses on estimating the uncertainty only about the odometry output (pure odometry), without considering long-term consistency issues \cite{costante2020uncertainty}. Thereby, neither of the methods utilize the compounded uncertainty to adequately balance the global constraint while the methods with consistency constraints require rigorous tuning of the weighting between loss terms with convergence issues directly related to unprincipled weighting approaches. Meanwhile, common architectures for odometry consist of recurrent modules. Therefore, the ability of the network in performing backpropagation through time alongside the lack of an appropriate framework for loss tuning motivates us to develop a principled approach to a consistency based loss term without stability issues.

\begin{figure}[t]
  \includegraphics[width=0.995\linewidth]{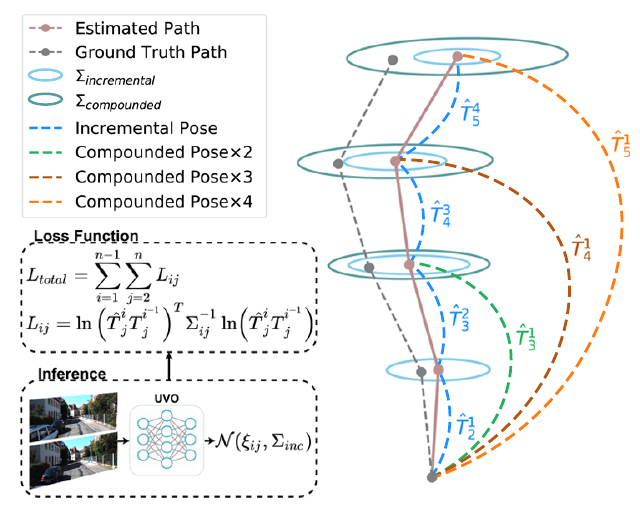}
  \caption{An overview of the proposed method. The incremental and compounded uncertainties are shown by projecting the covariance ellipsoid onto a 2-D plane. Consistency based loss terms are formed using the propagated covariance matrices. Note that the overlapping windows of integration for each window size are not shown for clarity.}
  \label{fig:figure1}
\end{figure}

In this paper, we propose to compound the uncertainties estimated by the network at each iteration of the algorithm and use the resulting covariance matrices to impose a consistency based constraint on the network. This method allows us to provide an adaptive method of weighting the incremental loss against the integrated loss while also allowing the network to tune the importance of motion on each axis. An overview of our approach is shown in Fig. \ref{fig:figure1}. We implement our proposed method in a Visual Odometry (VO) setting where we first infer a probability distribution over the SE(3) pose output of the network using a pair of input images. Then, we use the incremental outputs alongside the compounded pose and uncertainty values to form the proposed loss function. We quantitatively and qualitatively compare our results against the current classical and learning-based state of the art (SOTA) methods while outperforming recent work on both categories. Thereafter, we evaluate the uncertainties derived by the network and provide an in-depth analysis of the effects of the resulting covariance matrices as loss weighting medium. Finally, we utilize a loop detection algorithm to demonstrate the effectiveness of estimated odometry uncertainties in a pose-graph setup. To the best of our knowledge, propagation of uncertainty has not been proposed as a part of the loss function of an odometry network and this is the first approach that takes accumulation of uncertainty into account in such a setting. Briefly, our contributions are as follows:
\begin{itemize}
    \item We propose a consistency based loss function for deep odometry algorithms based on uncertainty compounding and provide quantitative and qualitative comparisons while outperforming the SOTA,
    \item Rigorous analysis on the effect of the compounded term on the loss value is provided,
    \item We embed our method into a pose-graph alongside a loop closure detection algorithm to showcase the importance of the uncertainties estimated by the network in a hybrid localization system.
\end{itemize}
This paper is structured as follows. In Section \ref{sec:related} related works from the literature are discussed and the difference between recent methods and ours is delineated. The details of our proposed approach is provided in Section \ref{sec:ourapproach} alongside the implementation method and the architecture of our network. Finally in Section \ref{sec:results}, we report the results of our experiments and provide qualitative and quantitative analysis of our network alongside that of SOTA.

\section{RELATED WORKS}
\label{sec:related}
From an algorithmic perspective, uncertainty in odometry has been proposed in standalone deep learning \cite{wang2018end} and hybrid algorithms \cite{liu2018deep}. Regardless of the uncertainty quantification formulation, deep learning based methods commonly take a maximum-likelihood approach to bypass the need for labels for the covariance matrix at each step. Alternatively in hybrid cases, deep learning based uncertainty estimation is utilized to estimate the error distribution of classical VO systems or used in conjunction with various filters such as the extended Kalman filter in a tightly coupled state estimation scenario \cite{li2020towards}. We briefly discuss both categories in this section.

DeepVO \cite{wang2017deepvo}, was the first work to formulate VO in an end-to-end fashion. This network computes the odometry without considering the long-term consistency issues and uncertainty surrounding the estimated pose. This work was later extended to ESP-VO \cite{wang2018end} to account for the frame-to-frame uncertainties of the output poses. However, this work does not take the increase in the uncertainty of poses into account while imposing a global constraint. In contrast, we propagate the uncertainties estimated at each iteration to account for the uncertainties beyond a single step.

CL-VO \cite{saputra2019learning}, proposes to integrate the odometry estimates to create a consistency-based loss term. This work does not associate uncertainty with the output poses. Due to the lack of adaptive weighting parameters for the loss terms, \cite{saputra2019learning} requires manual tuning of the loss functions. Moreover, the proposed loss function in CLVO uses a handcrafted scheduling system to determine when to include the long-term error in the overall loss. In our work, apart from associating uncertainty with each output, we also propagate the uncertainty to weigh the global loss term, eliminating the need for loss tuning or scheduling.

UA-VO \cite{costante2020uncertainty}, uses a conventional CNN-LSTM architecture to estimate the odometry poses alongside their uncertainty. This work extends the previous works by including the epistemic uncertainty of the network during inference through calculation of the predictive uncertainty. UA-VO does not take into account the long-term consistency issues and has no loss terms that minimize the output errors beyond frame-to-frame deviations.

Deep Inference for Covariance Estimation (DICE) \cite{liu2018deep}, estimates the error distribution of an arbitrary classical odometry method using a CNN that takes as input a single image from the pair that was passed to the classical VO pipeline. Deeper-Dice \cite{maio2020ddice}, extends this method by adding the corrections from the network estimates to the VO output before modeling their distribution to account for the biases of the VO outputs. Our method does not require a separate classical pipeline to estimate the odometry and we infer the odometry itself alongside the covariance matrix using a single network.

\section{Proposed Approach and Architecture}
\label{sec:ourapproach}
Odometry is defined as estimation of the incremental movement of a device where at each iteration, the change in rotation and position of the device is derived. To lower the number of outputs from the network, pose estimates are commonly represented using formulations other than SE(3) matrices such as se(3) lie algebra vectors. Therefore, to compute the predicted global pose of the device, the odometry estimates may first be converted to the corresponding SE(3) matrix representation and integration can then be performed as follows

\begin{equation}
    T^0_n = T^0_1\otimes T^1_2\otimes T^2_3\otimes ...\otimes T^{n-1}_n
    \label{eq:odometryintegration}
\end{equation}

Where $T^{i-1}_{i}$, represents the transformation matrix from frame $i-1$ to frame $i$. In the rest of this section, we will first associate uncertainty with each output of the network. Then, the uncertainty compounding formulation will be provided and our loss function will be proposed. Finally, the uncertainty quantification formulation using parametric methods such as neural networks will be discussed and the architectural details of the network will be provided.

\subsection{Incremental Pose Uncertainty}
\label{sec:posepdf}
There are several works on the association of uncertainty with pose vectors \cite{chirikjian2011stochastic, su1991prop, barfoot2014associating}. In this paper, we adopt the vector space of the SE(3) group as the pose output of the network and define a PDF on the se(3) vectorspace which in turn allows us to induce uncertainty on the SE(3) matrices through the exponential mapping. To this end, we use noisy perturbations \cite{barfoot2014associating} to associate uncertainty with SE(3) matrices as follows

\begin{equation}
    T = \boldsymbol{e}^{\boldsymbol{\xi}} \Bar{T}
    \label{eq:perturbse3}
\end{equation}

in which $\boldsymbol{\xi}$ represents the noisy perturbation and is defined as a zero-mean Gaussian with covariance matrix $\mathbf{\Sigma}$ as below

\begin{equation}
    p(\boldsymbol{\xi})=\mathcal{N}(\mathbf{0}, \mathbf{\Sigma}), \mathbf{\Sigma}\in R^{6\times 6}
    \label{eq:perturbgaussian}
\end{equation}

The PDF over the vectorspace can then be defined as \cite{barfoot2014associating}

\begin{equation}
\int_{\mathbb{R}^{6}} p(\boldsymbol{\xi}) d \boldsymbol{\xi}=\int_{\mathbb{R}^{6}} \eta \boldsymbol{e}^{(-\frac{1}{2} \boldsymbol{\xi}^{T} \boldsymbol{\Sigma}^{-1} \boldsymbol{\xi})} d \boldsymbol{\xi}=1
\label{eq:perturbpdf}
\end{equation}

Where $\eta$ represents the normalization factor and is defined as $\eta=\frac{1}{\sqrt{(2 \pi)^{6} \operatorname{det}(\mathbf{\Sigma})}}$.

\subsection{Uncertainty Compounding}
To integrate the odometry output from the network while propagating the incremental uncertainty, we use the definition from (\ref{eq:perturbse3}) as follows

\begin{equation}
    \boldsymbol{e}^{\boldsymbol{\xi}_{02}}T^{i-2}_{i} = \boldsymbol{e}^{\boldsymbol{\xi}_{12}} T^{i-2}_{i-1}~\boldsymbol{e}^{\boldsymbol{\xi}_{01}} T^{i-1}_{i}
    \label{eq:compoundingtoy}
\end{equation}

Where $T^{i-2}_{i}$ represents the mean global transformation matrix with the compounded uncertainty in the form of a noisy perturbation represented by $\boldsymbol{e}^{\boldsymbol{\xi_{02}}}$. Moreover, $\boldsymbol{e}^{\boldsymbol{\xi_{01}}} T^{i-1}_{i}$ and $\boldsymbol{e}^{\boldsymbol{\xi_{12}}} T^{i-2}_{i-1}$ represent the consecutive outputs from the network in 2 iterations over a trajectory.
To derive the formulation for calculating $\boldsymbol{e}^{\boldsymbol{\xi_{02}}}$, we use the Baker-Campbell-Hausdorff (BCH) formula following \cite{barfoot2017state}, to which we refer the reader for a full interpretation. 

The BCH formula is an infinite series that provides a solution to the multiplication of the exponential of two elements from the vectorspace of a lie group as shown in the equation below

\begin{equation}
Z=log(\boldsymbol{e}^{X}\boldsymbol{e}^{Y}) 
\label{eq:BCHappearence}
\end{equation}

where $X$ and $Y$ belong to the lie algebra of a lie group. The solution to this equation is as follows \cite{Klarsfeld1989bch}

\begin{equation}
\begin{aligned}
Z &=X+Y \\
&+\frac{1}{2}[X, Y] \\
&+\frac{1}{12}([X,[X, Y]]+[Y,[Y, X]]) \\
&+\frac{1}{48}([Y,[X,[Y, X]]]+[X,[Y,[Y, X]]]) \\
&+\cdots
\end{aligned}
\label{eq:BCHgeneralsolution}
\end{equation}

where $[X, Y]=XY-YX$ is the Lie bracket. To solve (\ref{eq:compoundingtoy}) for $\boldsymbol{\xi_{02}}$, we first need to manipulate the right hand side of (\ref{eq:compoundingtoy}) to be similar to that of (\ref{eq:BCHappearence}). By moving the perturbation factors to the left hand side of $T^{i-2}_{i-1}$ we have

\begin{equation}
\boldsymbol{e}^{(\boldsymbol{\xi_{02}}^{\wedge})}=\boldsymbol{e}^{(\boldsymbol{\xi_{12}}^{\wedge})} \boldsymbol{e}^ {(\overline{\mathcal{T}}^{i-2}_{i-1} \boldsymbol{\xi_{01}})^{\wedge}}
\label{eq:compoundingBCH}
\end{equation}

in which $\overline{\mathcal{T}}^{i-2}_{i-1}$ is the adjoint of the matrix $\boldsymbol{T}^{i-2}_{i-1}$ and the wedge ($^\wedge$) operator is defined as below

\begin{equation}
\boldsymbol{\xi}^\wedge = 
\begin{bmatrix}
\boldsymbol{\rho} \\
\boldsymbol{\phi} \\
\end{bmatrix}^\wedge = 
\begin{bmatrix}
\boldsymbol{\phi}^\wedge & \boldsymbol{\rho} \\
\mathbf{0}_{1\times 3}& 0\\
\end{bmatrix}, \boldsymbol{\xi} \in \mathop{\mathbb{R}^6}
\label{eq:wedgeoperation}
\end{equation}

By using the BCH formula on (\ref{eq:compoundingBCH}) while noting that $\mathop{\mathbb{E}}[\boldsymbol{\xi}_{ij}]=0$ for any $i$ and $j$, we can derive the covariance matrix of the compounded uncertainty as follows \cite{barfoot2014associating}

\begin{equation}
\begin{aligned}
\boldsymbol{\Sigma}_{02} = \mathop{\mathbb{E}}[\boldsymbol{\xi}_{02}\boldsymbol{\xi}^{T}_{02}] &= \mathop{\mathbb{E}}[\boldsymbol{\xi}_{12}\boldsymbol{\xi}^T_{12}+\boldsymbol{\xi}^\prime_{01}\boldsymbol{\xi}^{\prime^T}_{01}\\
+\frac{1}{12}&((\boldsymbol{\xi}^\curlywedge_{12}\boldsymbol{\xi}^\curlywedge_{12})(\boldsymbol{\xi}^\prime_{01}\boldsymbol{\xi}^{\prime^T}_{01})\\
&+(\boldsymbol{\xi}^\prime_{01}\boldsymbol{\xi}^{\prime^T}_{01})(\boldsymbol{\xi}^\curlywedge_{12}\boldsymbol{\xi}^\curlywedge_{12})^T\\
&+(\boldsymbol{\xi}^{\prime^\curlywedge}_{01}\boldsymbol{\xi}^{\prime^\curlywedge}_{01})(\boldsymbol{\xi}_{12}\boldsymbol{\xi}^T_{12})\\
&+(\boldsymbol{\xi}_{12}\boldsymbol{\xi}^T_{12})(\boldsymbol{\xi}^{\prime^\curlywedge}_{01}\boldsymbol{\xi}^{\prime^\curlywedge}_{01})^T)\\
+\frac{1}{4}&(\boldsymbol{\xi}^\curlywedge_{12}(\boldsymbol{\xi}^\prime_{01}\boldsymbol{\xi}^{\prime^T}_{01})\boldsymbol{\xi}^{\curlywedge^T}_{12})]
\label{eq:compoundingsigma}
\end{aligned}
\end{equation}

where $\boldsymbol{\Sigma}_{02}$ is the compounded covariance matrix and $\boldsymbol{\xi}^\prime_{01}=\overline{\mathcal{T}}^{i-2}_{i-1} \boldsymbol{\xi}_{01}$. The curly wedge operation ($^\curlywedge$) is defined as

\begin{equation}
\boldsymbol{\xi}^\curlywedge = 
\begin{bmatrix}
\boldsymbol{\rho} \\
\boldsymbol{\phi} \\
\end{bmatrix}^\curlywedge = 
\begin{bmatrix}
\boldsymbol{\phi}^\wedge & \boldsymbol{\rho}^\wedge \\
\mathbf{0}_{3\times3} & \boldsymbol{\phi}^\wedge\\
\end{bmatrix},~\boldsymbol{\rho}, \boldsymbol{\phi} \in \mathop{\mathbb{R}^3}
\label{eq:curlywedge}
\end{equation}

Using (\ref{eq:curlywedge}) and noting $\lambda_1^\wedge\lambda_2^\wedge=-(\lambda_1^T\lambda_2)1+\lambda_2\lambda_1^T$, (\ref{eq:compoundingsigma}) may be broken down to

\begin{align}
\mathop{\mathbb{E}}[\boldsymbol{\xi}_{12}\boldsymbol{\xi}^T_{12}] &= \boldsymbol{\Sigma}_{12}
\label{eq:compoundingsigmasolution1}\\
\mathop{\mathbb{E}}[\boldsymbol{\xi}^\prime_{01}\boldsymbol{\xi}^{\prime^T}_{01}] &= \boldsymbol{\Sigma}_{01}^{\prime} = \overline{\mathcal{T}}^{i-2}_{i-1} \boldsymbol{\Sigma}_{01} \overline{\mathcal{T}}^{i-2}_{i-1}\\
\mathop{\mathbb{E}}[\boldsymbol{\xi}^\wedge_{12}\boldsymbol{\xi}^\wedge_{12}] &= 
\begin{bmatrix}
(\boldsymbol{\Sigma}^{12}_{\phi\phi})^* & (\boldsymbol{\Sigma}^{12}_{\rho\phi}+\boldsymbol{\Sigma}^{12^T}_{\rho\phi})^*\\
\mathbf{0}_{3\times 3} & \boldsymbol{\Sigma}^{12}_{\phi\phi}\\
\end{bmatrix} \\
\mathop{\mathbb{E}}[\boldsymbol{\xi}^{\prime^\wedge}_{01}\boldsymbol{\xi}^{\prime^\wedge}_{01}] &= 
\begin{bmatrix}
\boldsymbol{(\Sigma}_{\phi\phi}^{01^\prime})^* & (\boldsymbol{\Sigma}_{\rho\phi}^{01^\prime}+\boldsymbol{\Sigma}_{\rho\phi}^{01^{\prime^T}})^* \\
\mathbf{0}_{3\times 3} & \boldsymbol{\Sigma}_{\phi\phi}^{01^\prime}\\
\end{bmatrix}\\
\mathop{\mathbb{E}}[\boldsymbol{\xi}^\curlywedge_{12}(\boldsymbol{\xi}^\prime_{01}&\boldsymbol{\xi}^{\prime^T}_{01})\boldsymbol{\xi}^{\prime^T}_{12}] = \begin{bmatrix}
\mathcal{B}_{11} & \mathcal{B}_{12}\\
\mathcal{B}_{21} & \mathcal{B}_{22}\\
\end{bmatrix}\\
\begin{split}
\mathcal{B}_{11} &= (\boldsymbol{\Sigma}_{12}^{\phi\phi}, \boldsymbol{\Sigma}_{01}^{\rho\rho^\prime})^*+(\boldsymbol{\Sigma}_{12}^{\rho\phi^T}, \boldsymbol{\Sigma}_{01}^{\rho\phi^\prime})^*\\&+(\boldsymbol{\Sigma}_{12}^{\rho\phi}, \boldsymbol{\Sigma}_{01}^{\rho\phi^{\prime^T}})^*+(\boldsymbol{\Sigma}_{12}^{\rho\rho}, \boldsymbol{\Sigma}_{01}^{\phi\phi^\prime})^*
\end{split}\\
\begin{split}
\mathcal{B}_{12} &= (\boldsymbol{\Sigma}_{12}^{\phi\phi}, \boldsymbol{\Sigma}_{01}^{\rho\phi^{\prime^T}})^*+(\boldsymbol{\Sigma}_{12}^{\rho\phi^T}, \boldsymbol{\Sigma}_{01}^{\phi\phi^\prime})^*
\end{split}\\
\mathcal{B}_{21} &= \mathcal{B}_{01}^T\\
\begin{split}
\mathcal{B}_{22} &= (\boldsymbol{\Sigma}_{12}^{\phi\phi}, \boldsymbol{\Sigma}_{01}^{\phi\phi^{\prime}})^*
\end{split}
\label{eq:compoundingsigmasolution}
\end{align}

where $A^* = -tr(A)1 + A$ and $(A, B)^* = A^*B^*+(BA)^*$. Therefore, (\ref{eq:compoundingsigmasolution1})-(\ref{eq:compoundingsigmasolution}) can be used to calculate the compounded uncertainty while the mean value of the compounded pose may be found through $T^{i-2}_{i} =  T^{i-2}_{i-1}T^{i-1}_{i}$.

\subsection{Loss Function}
In this section, we will treat odometry as a multi-task learning problem and factorize a likelihood over the incremental outputs of the network and their integration to derive the loss that needs to be minimized. For the problem with two odometry outputs from the network, defined in (\ref{eq:compoundingtoy}), we have the following likelihood
\begin{equation}
\begin{split}
p(\boldsymbol{\xi}_{1}, \boldsymbol{\xi}_{2}, \boldsymbol{\xi}_{1:2}& \mid \mathbf{f}_{\boldsymbol{\theta}}(I_{1,2,3}))=\\ &p(\boldsymbol{\xi}_{1} \mid \mathbf{f}_{\boldsymbol{\theta}}(I_{1, 2})) \\\times~& p(\boldsymbol{\xi}_{2} \mid \mathbf{f}_{\boldsymbol{\theta}}(I_{2, 3})) \\\times~& p(\boldsymbol{\xi}_{1:2} \mid \mathbf{f}_{\boldsymbol{\theta}}(I_{1,2,3}))
\end{split}
\label{eq:factorization}
\end{equation}
Where $I_i$ represents the input frame at iteration $i$ and $f_\theta$ represents the function that takes $I_i$ as input and outputs $\boldsymbol{\xi_{i}}$ using parameters $\theta$. Moreover, $\boldsymbol{\xi}_{1}=\log(T^{i-2}_{i-1})$ and $\boldsymbol{\xi}_{2}=\log(T^{i-1}_{i})$ represent the consecutive estimates of the network based on the input frames $I_{1,2}$ and $I_{2,3}$, respectively, while $\boldsymbol{\xi}_{1:2}=\log(T^{i-2}_{i-1}T^{i-1}_{i})$ represents the lie algebra vector corresponding to the integrated pose. The negative log likelihood of (\ref{eq:factorization}) derives the objective that needs to be minimized
\begin{equation}
\begin{split}
-\log~p(&\boldsymbol{\xi}_{1}, \boldsymbol{\xi}_{2}, \boldsymbol{\xi}_{1:2} \mid \mathbf{f}_{\boldsymbol{\theta}}(I_{1,2,3}))=\\ 
& \log(e^{\boldsymbol{\xi}_{1}}\mathbf{T}_{i-1}^{i-2^{-1}})\boldsymbol{\Sigma}_{12}^{-1}\log(e^{\boldsymbol{\xi}_{1}}\mathbf{T}_{i-1}^{i-2^{-1}})^T\\
+& \log(e^{\boldsymbol{\xi}_{2}}\mathbf{T}_{i}^{i-1^{-1}})\boldsymbol{\Sigma}_{01}^{-1}\log(e^{\boldsymbol{\xi}_{2}}\mathbf{T}_{i}^{i-1^{-1}})^T\\
+& \log(e^{\boldsymbol{\xi}_{1:2}}\mathbf{T}_{i}^{i-2^{-1}})\boldsymbol{\Sigma}_{02}^{-1}\log(e^{\boldsymbol{\xi}_{1:2}}\mathbf{T}_{i}^{i-2^{-1}})^T\\
+& \log(|\boldsymbol{\Sigma}_{12}|)+\log(|\boldsymbol{\Sigma}_{01}|)+\log(|\boldsymbol{\Sigma}_{02}|)
\end{split}
\label{eq:negloglikelihood}
\end{equation}

where $\mathbf{T}$ represents the ground truth pose and the first three terms on the right hand side represent the geodesic distance between the estimated and ground truth poses weighted by the covariance matrix estimated by the network itself.

The overall loss is boiled down to two terms defined by
\begin{equation}
\begin{split}
\mathcal{L} &= \mathcal{L_{\text{incremental}}}+\mathcal{L_{\text{composed}}}\\
\end{split}
\label{eq:negloglikelihood}
\end{equation}

The incremental loss term in the right-hand side of (\ref{eq:negloglikelihood}) minimizes the error of the estimated frame-to-frame motion, while the composed loss minimizes the deviation of the estimates from the global path over a window of outputs. Moreover, odometry loss terms are weighted by the covariance matrix estimated at that iteration and the global losses are weighted by the compounded matrix that is the result of (\ref{eq:compoundingsigma}). Therefore, 
in the case of frame to frame loss, if the network is not able to estimate the output accurately, it can increase the uncertainty output to lower the amount of loss. On the other hand, the last three terms in  (\ref{eq:negloglikelihood}) act as regularizers and punish large uncertainties to create an overall balance. In case of the global loss term, the loss values are weighted by the compounded covariance matrix. This means that if at a certain iteration along the trajectory, a pair of input frames result in a peak over the pose uncertainty (the network was not able to estimate the output accurately) the propagated uncertainty will substantially increase during the compounding process and the integrated loss will be adaptively weighted. Therefore, uncertainty quantification allows us to weigh the motion on each axis while also providing an adequate way to balance the short-term and long-term losses against each other.

\begin{figure*}[t]
   \hspace{0.4cm}
   \includegraphics[width=0.93\linewidth]{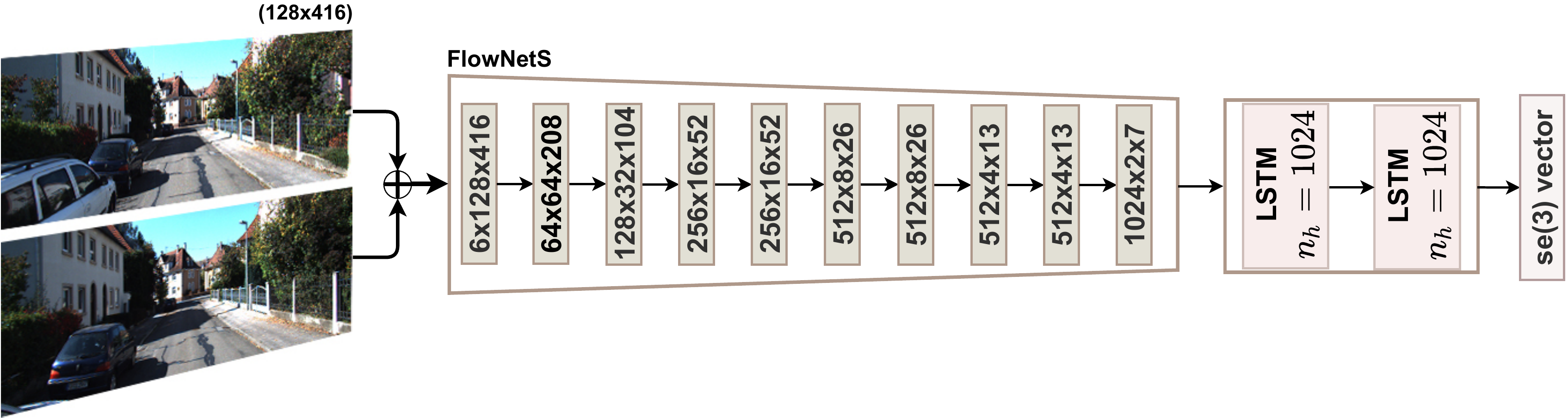}
   \caption{Architecture of our VO network. The images are resized, concatenated along the channel dimension and passed to the network for processing.}
   \label{fig:architecture}
\end{figure*}

\subsection{Implementation Details}
In this section we provide the details of the uncertainty quantification algorithm and delineate the architecture of our neural network.

\subsubsection{Network Architecture}
We use a CNN-LSTM architecture to derive a spatio-temporal model of the consecutive inputs. As can be seen in Fig. \ref{fig:architecture} we use a 9 layer CNN to extract the visual features from a pair of input frames. To achieve faster and more efficient training, we initialize the CNN with weights from an optical flow network \cite{ilg2017flownet}.
The visual features are then converted into a vector using global average pooling. This layer averages the spatial features and outputs a vector with the same length as the depth of the input feature map. The averaged features are then passed through two layers of Long-Short Term Memory networks to model the visual features temporally. Thereafter, two fully connected layers (not shown in Fig. \ref{fig:architecture}) are used to estimate the output pose and uncertainty. In particular, we infer the output pose alongside the diagonal covariance matrices in the form of a vector with a length of 12 from which 6 correspond to the incremental pose represented using se(3) vectors and the rest represent the uncertainty over each motion axis. 

\subsubsection{Uncertainty Quantification}
The uncertainty quantification formulation should be constrained in such a way that the resulting matrix would be semi-positive definite. To this end, We process the 6 uncertainty outputs into diagonal elements of the covariance matrix through $\sigma_i^2~=~\exp{(s_i)}$ where $s_i=\log\sigma^2$ is estimated by the network. To calculate $\log{|\boldsymbol{\Sigma}|}$ from (\ref{eq:negloglikelihood}) the following equation may be used

\begin{equation}
\log{|\boldsymbol{\Sigma}|}=\log(\prod_{i=1}^{n=6}\sigma^{2}_{i}) = \sum_{i=1}^{n=6}\log(\sigma^2_i) = \sum_{i=1}^{n=6}(\text{s}_i)
\label{eq:quantifiationf2f}
\end{equation}

On the other hand, the term $\log{|\boldsymbol{\Sigma}|}$ corresponding to the compounded loss term in (\ref{eq:negloglikelihood}) is no longer diagonal due to the compounding, and (\ref{eq:quantifiationf2f}) cannot be used to calculate this term. To this end, we take the Cholesky factorization of the estimated covariance matrix and calculate $\log{|\boldsymbol{\Sigma}|}$ as follows

\begin{equation}
\log{|\boldsymbol{\Sigma}|}=\log(|LL^T|) = 2\log(|L|) = 2\sum_{i=1}^{n=6}(\log{L_{i i}})
\label{eq:quantifiationcompounded}
\end{equation}

where $L$ is the lower triangular matrix resulting from Cholesky factorization of $\boldsymbol{\Sigma}$.

\section{Experiments and Analysis}
\label{sec:results}
We perform all the experiments on an NVIDIA P100 GPU using PyTorch and PyTorch lightning. While training, we use short segments of the training sequences with lengths of 32. The windows over which output poses are compounded have a maximum length of 5 while a batch size of 16 is used during training. Moreover, we have open--sourced our code for reproducibility  purposes\footnote{The code will be available upon acceptance}. In the following, we discuss the dataset used for all our analyses alongside the approaches against which we compare our method.

\begin{table*}[!t]
    \centering
    \caption{Quantitative Analysis}
    \begin{tabular}{c|cc|cccc|c} \toprule
        \multirow{2}{*}{Sequence} & {DSO \cite{engel2017direct}} & {ORB-SLAM2 \cite{mur2017orb}} & {DeepVO \cite{wang2017deepvo}} & {CLVO \cite{saputra2019learning}} & {ESPVO \cite{wang2018end}} & {UA-VO \cite{costante2020uncertainty}} & {UVO (ours)} \\
        & {t(\%)/r($^\circ$)} & {t(\%)/r($^\circ$)} & {t(\%)/r($^\circ$)} & {t(\%)/r($^\circ$)} & {t(\%)/r($^\circ$)} & {t(\%)/r($^\circ$)} & {t(\%)/r($^\circ$)}\\\midrule
        08 & 49.2/\textbf{0.44} & 57.2/0.46 & 9.06/2.64 & 8.84/2.88 & 11.60/4.27 & 9.68(7.91)/3.82(2.76) & \textbf{5.12}/\textbf{1.35}\\
        09 & 67.6/\textbf{0.52} & 72.0/0.84 & 10.6/4.21 & 8.83/3.54 & 11.28/3.22  & 10.2(11.9)/4.29(3.15) & \textbf{8.31}/\textbf{2.63} \\
        10 & 77.3/1.43 & 83.0/\textbf{0.51} & 15.8/4.14 & 14.5/3.90 & 12.66/4.32  & 11.1(10.3)/3.86(3.49) & \textbf{10.5}/\textbf{2.91}\\ \midrule
        Avg. & 64.7/0.80 & 70.7/\textbf{0.60} & 11.8/3.66 & 10.72/3.44  & 11.85/3.94   & 9.95(10.0)/3.93(3.13) & \textbf{7.98}/\textbf{2.30} \\ \bottomrule
    \end{tabular}
    \label{tab:pathquantiative}
\end{table*}

\subsection{Dataset and Evaluation}
We use the KITTI odometry dataset to perform our experiments. This dataset consists of 22 sequences of driving a car in urban and residential areas. The first 11 sequences consist of stereo images alongside the ground truth pose while the remaining sequences are provided without ground truth. We use sequences 00-07 to train and validate our network and perform tests using sequences 08-10. To quantitatively evaluate our network we use the KITTI odometry benchmark \cite{geiger2012we}, where the relative translation and rotation errors of output poses are computed over segments with lengths of 100m-800m. For training and inference, we resize the images from the KITTI dataset to $128\times416$ pixels.

\subsection{Comparisons}
We compare our results against both classical and deep learning based odometry methods on the KITTI dataset. To compare with the classical methods we chose DSO \cite{engel2017direct}, a SOTA direct odometry approach and the monocular variant of ORB-SLAM2 \cite{mur2017orb} as a well-known SOTA indirect odometry method. To compare against deep learning based approaches, we chose UA-VO \cite{costante2020uncertainty}, ESP-VO \cite{wang2018end}, DeepVO \cite{wang2017deepvo} and CLVO \cite{saputra2019learning}. UA-VO is the current SOTA for uncertainty based odometry approaches. The loss function proposed in this method does not include a global term that would take long-term deviations into account. ESP-VO and CLVO both include a compounding term in their loss function but do not make use of uncertainty to weigh the losses in a principled way. Finally, DeepVO is the SOTA odometry method that does not make use of uncertainty nor a global loss term.

\subsection{Quantitative Analysis}
The quantitative analysis of our method is provided in Table \ref{tab:pathquantiative} alongside the competing classical and deep learning based approaches. The results for the SOTA deep learning based method termed UA-VO are reported from \cite{costante2020uncertainty}. Furthermore, The values inside the parentheses represent the results of our re-implementation of UA-VO. Due to a lack of open-source code for DeepVO, CLVO and ESPVO, we implemented these methods based on \cite{wang2017deepvo, wang2018end, saputra2019learning}. 

When compared to deep learning based approaches, it can be seen that our method achieves a significantly higher accuracy both in terms of individual sequences and the overall mean. In particular, UVO obtains a 19.8\% increase in translation and 41.5\% increase in rotation accuracy over UA-VO. Among the other deep learning based methods, our method achieves an increase of 32.4\% over translation and 37.1\% over rotation accuracy compared to DeepVO which shows the benefits of using uncertainty-based losses alongside the proposed compositional loss term. Although CLVO does include a compositional loss term, the lack of adequate weighting results in a diminished accuracy compared to our approach. On the other hand, even though ESPVO does associate uncertainty with frame-to-frame outputs, the lack of such a weighting mechanism on the integrated poses degrades the performance of this network.

When compared to classical approaches, it can be seen that our approach consistently outperforms both DSO and ORBSLAM2 in terms of translation accuracy while the classical methods achieve higher accuracy in terms of rotation. In particular, UVO achieves an 8-fold increase in translation accuracy compared to ORBSLAM2 while this classical approach obtains a 4-fold increase in rotation accuracy compared to UVO. This shows one of the main downsides of classical monocular VO approaches, namely the problem of absolute scale recovery, that deep learning based methods solve through supervised learning.

\begin{figure}[t]%
    \centering
    \includegraphics[width=0.84\linewidth]{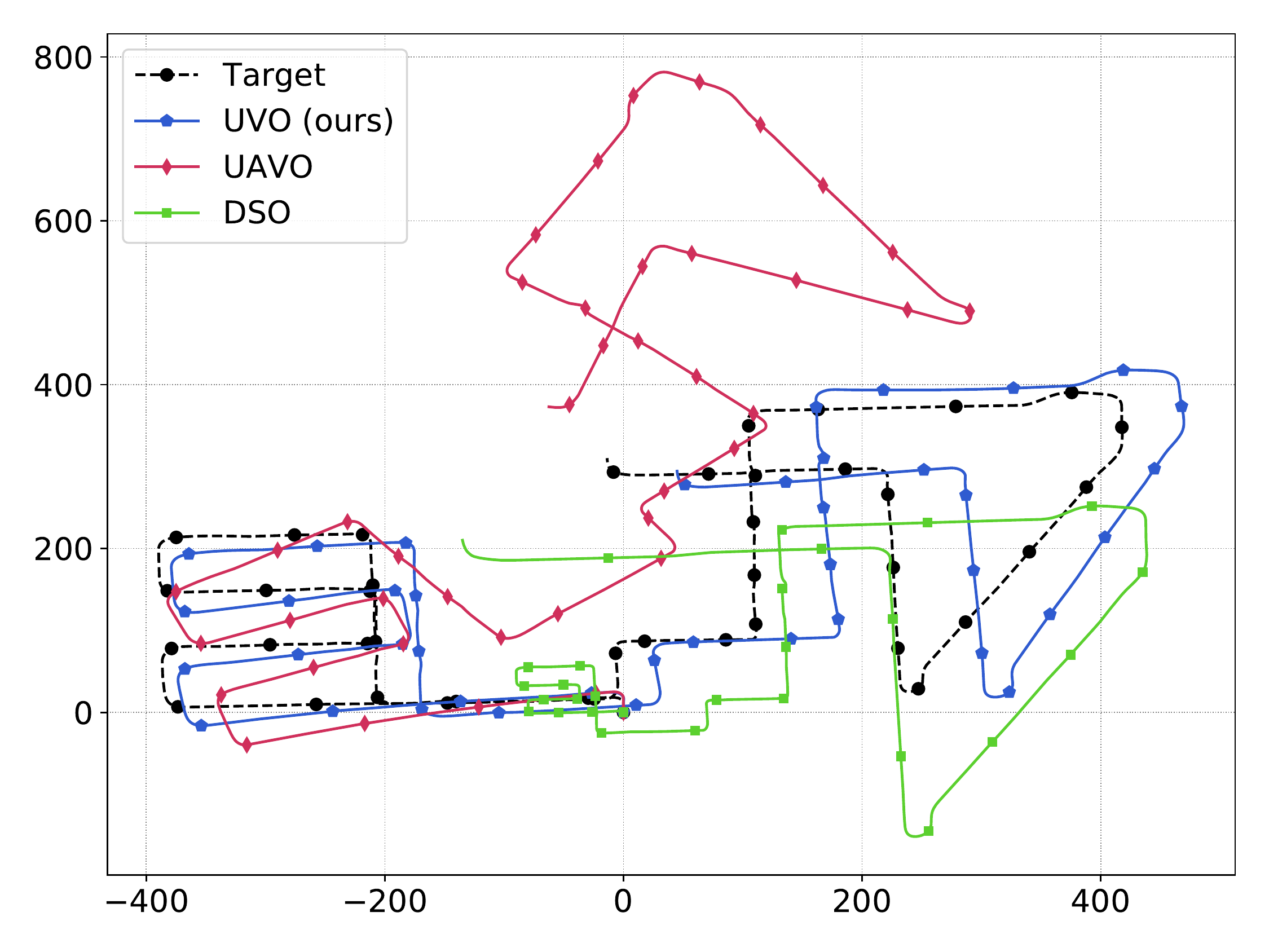}%
    \caption{Qualitative results on a test set of the KITTI dataset}
    \label{fig:qualitative}
    \vspace{-0.1cm}
\end{figure}

\subsection{Qualitative Analysis}
The qualitative result of our network in the form of the global path is presented in Fig. \ref{fig:qualitative} for test sequence 8 of the KITTI dataset. Based on the results from Fig. \ref{fig:qualitative}, our network is able to track the ground truth path more accurately compared to both deep learning based and classical approaches. Moreover, based on the first 300 meters of the trajectory, our network is able to maintain a low drift for a significantlsy larger distance compared to UA-VO. As mentioned in the previous section, due the unobservability of the absolute scale in classical odometry methods, the path for DSO in Fig. \ref{fig:qualitative} is scale-corrected. On the other hand, our method is able to estimate the absolute scale since this parameter is implicitly modeled during the training.

\begin{table}[!t]
    \centering
    \caption{Quantitative Evaluation of Uncertainty}
    \begin{tabular}{c|cc|cc} \toprule
        \multirow{2}{*}{Seq.} & \multicolumn{2}{c}{Fixed UI (Val.)} & \multicolumn{2}{c}{UVO (ours)} \\
               & OR(\%) & UI & OR(\%) & mUI \\ \midrule
        $\rho_x$ & 18.32 & 0.0086 & 2.35 & 0.0303 \\
        $\rho_y$ & 21.08 & 0.0059 & 3.86 & 0.0180 \\
        $\rho_z$ & 4.200 & 0.1212 & 2.24 & 0.2985 \\ \midrule
        $\phi_x$ & 20.12 & 0.0006 & 0.26 & 0.0032 \\
        $\phi_y$ & 18.37 & 0.0010 & 0.25 & 0.0077 \\
        $\phi_z$ & 16.80 & 0.0011 & 0.23 & 0.0050 \\ \bottomrule
    \end{tabular}
    \label{tab:uncertaintyevaluation}
\end{table}
\subsection{Uncertainty Evaluation}
To evaluate the uncertainty outputs, we calculate the percentage (OR\%) of samples in the KITTI dataset that fall out of the range of the distribution predicted by the network for each input. Ideally, we want this value to be close to zero meaning that the distribution predicted by the network contains the true value of the output. As a baseline, we use a validation set from the KITTI dataset and derive a fixed uncertainty interval based on the error of the network outputs on this validation set (assumption of homoscedasticity on the input noise). The results of this analysis are provided in Table \ref{tab:uncertaintyevaluation}. It can be seen that the distribution predicted by the network adequately covers the range that the true value of the outputs resides in, and on average, only 1.53\% of the true values fall out of the predicted range. Moreover, the mean uncertainty interval (UI) predicted by the network for the most prominent motion axis (axis representing vehicle's forward motion) is about 0.3 meters which is reasonable when compared to the mean displacement of the vehicle on this axis over the training distribution which is between 1m-3m over different sequences. Moreover, the OR for rotation on all axes is lower than that of translation. This is due to the lower amount of rotation experienced in the KITTI dataset relative to displacement. The UI for the displacement on other axes is 0.03 and 0.02 meters which are sensible due to the lower amount of motion that vehicles experience on lateral and vertical axes. On the other hand, a fixed UI results in a mean OR value of 16.48\%. This value alongside the UI shows that even though the UI was derived based on a validation set, it consistently results in overconfident intervals over all axes bringing about critical safety issues.

\begin{figure}[t]
\hspace{-0.28cm}
\subfigure[Translation weighting]{\includegraphics[width=1.755in]{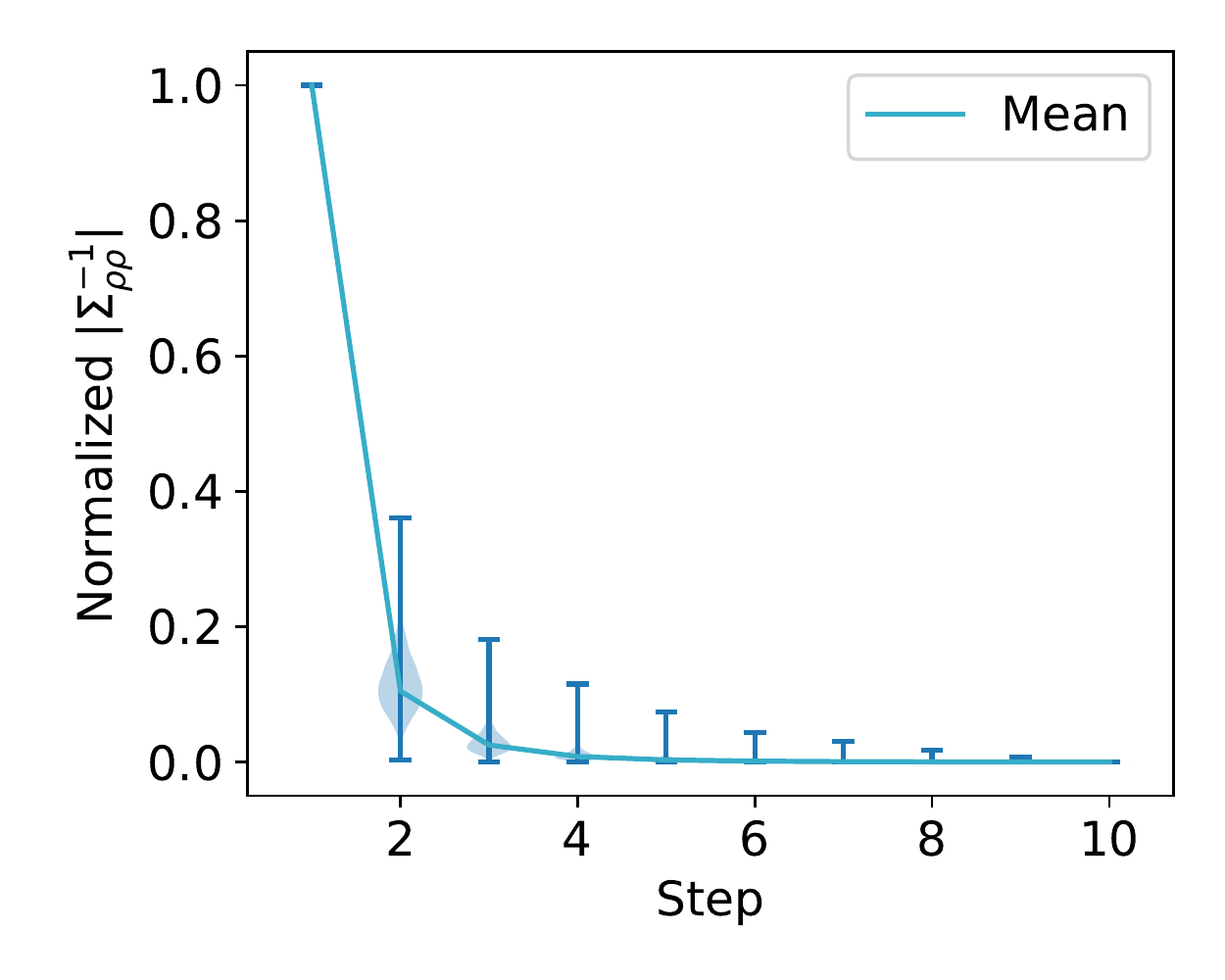}\label{fig:translationviolin}}
\hspace{-0.45cm}
\subfigure[Rotation Weighting]{\includegraphics[width=1.755in]{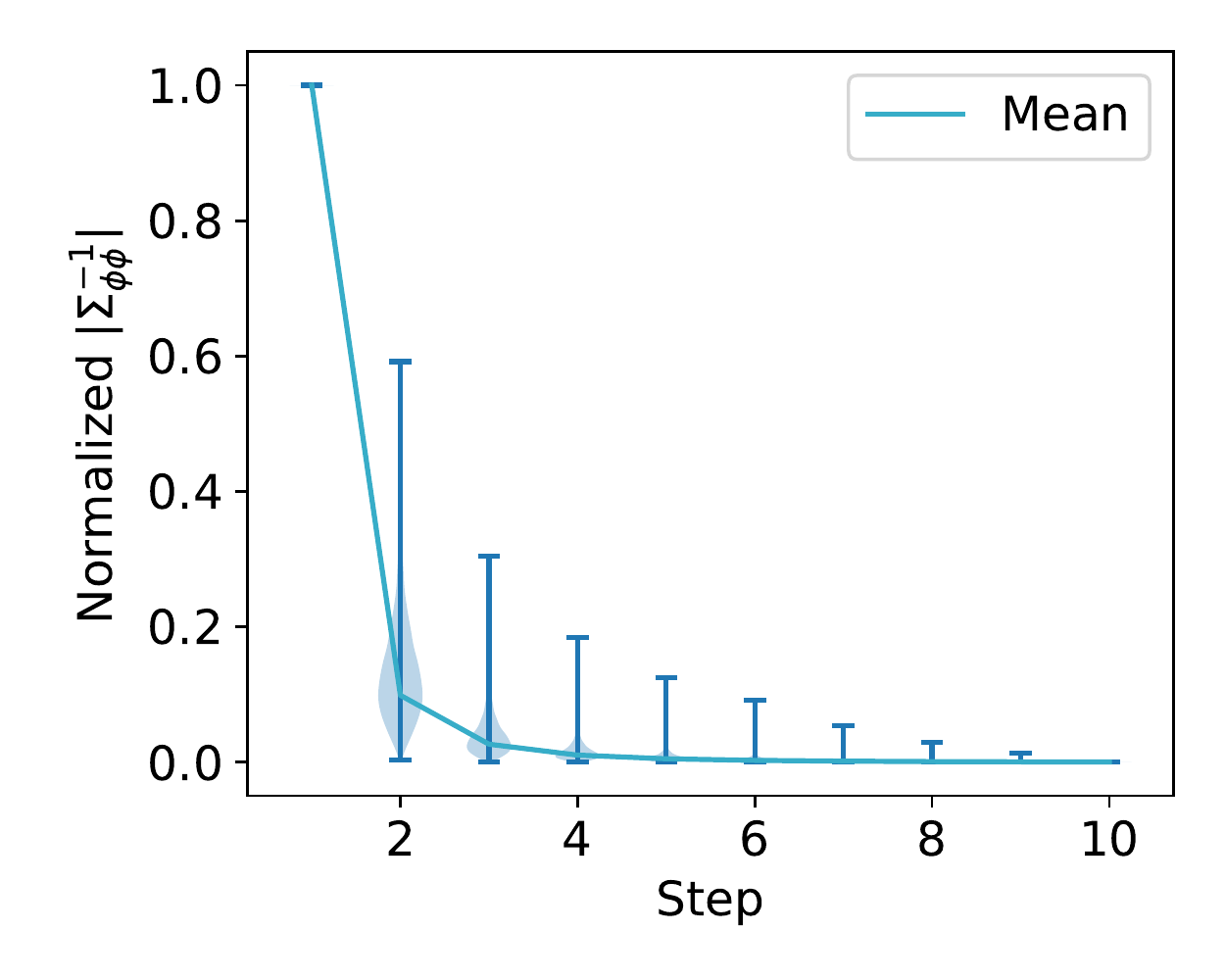}\label{fig:rotationviolin}}
\hfill
\vfill
\vspace{-0.35cm}
\hfill
\subfigure[Normalized compounded loss]{\raisebox{1mm}{\includegraphics[width=1.69in]{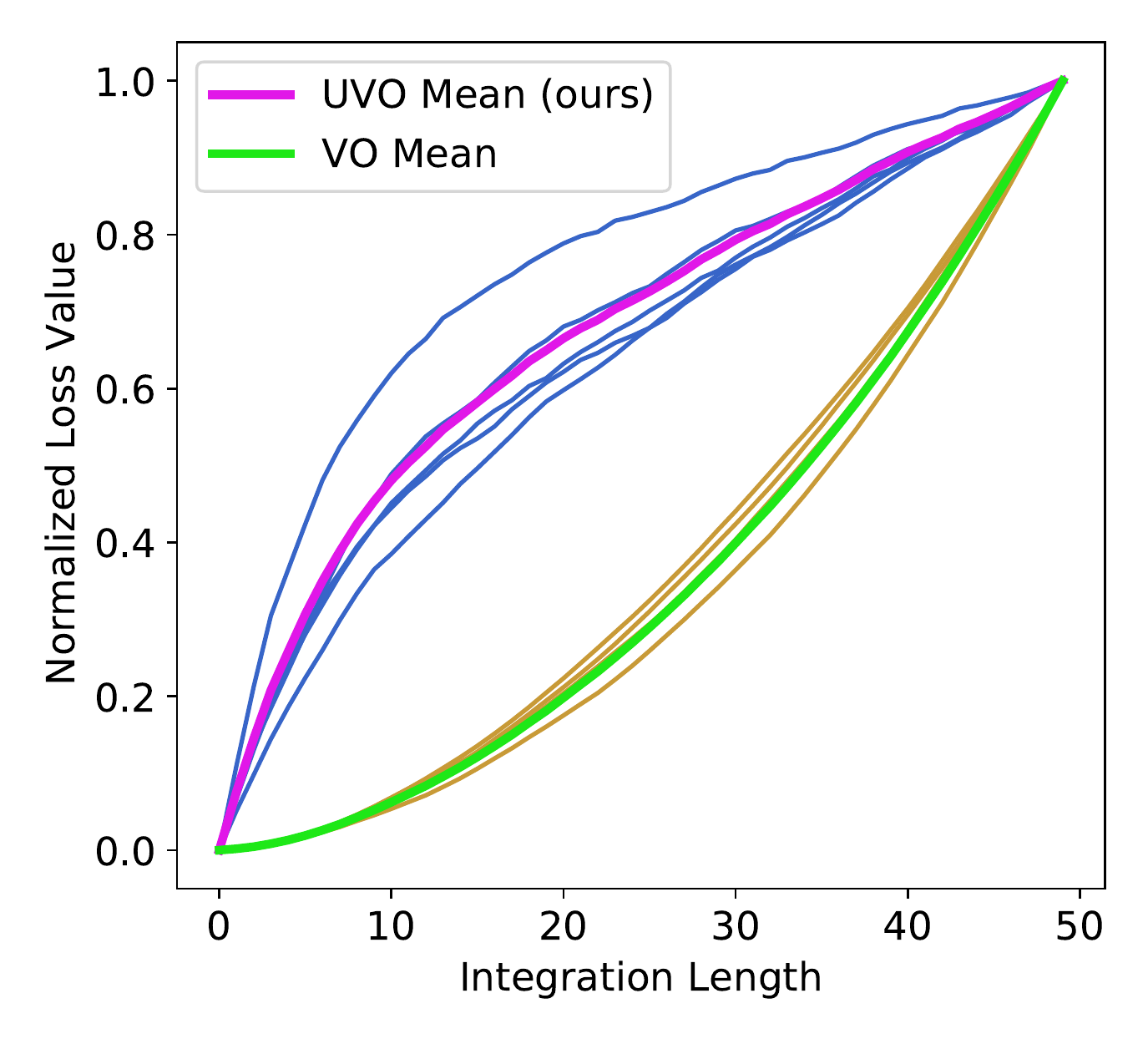}}\label{fig:lossvalue}}
\hspace{-0.4cm}
\subfigure[Normalized frame-to-frame loss]{\includegraphics[width=1.72in]{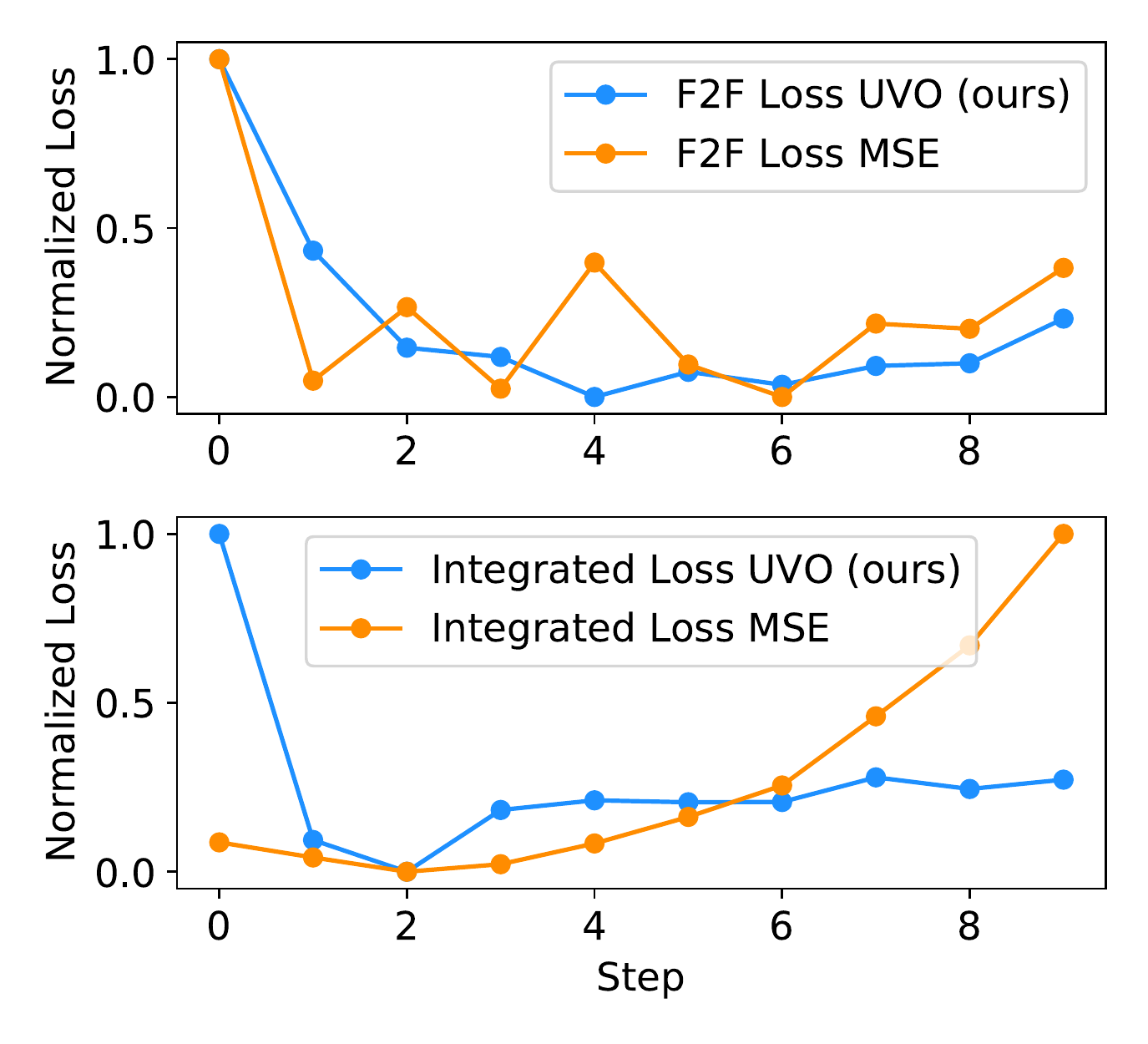}\label{fig:casestudy}}
\hfill
\caption{Effect of weighting on the loss values over the KITTI dataset}
\label{fig:volossvalue}
\vspace{-0.2cm}
\end{figure}
\subsection{Weighting Analysis}
In this section, we visualize the weighting derived by the network for each of the loss terms in (\ref{eq:negloglikelihood}). Fig. \ref{fig:translationviolin} and Fig. \ref{fig:rotationviolin} represent the normalized $|\boldsymbol{\Sigma}^{-1}|$ values for the translation and rotation sections of the covariance matrix, respectively. The X-axis of the two figures shows the number of outputs that have been compounded. Based on these two figures, the compounding of the covariance matrices induces exponentially decaying weighting terms for both translation and rotation as the number of steps increases. The direct effect of this approach to weighting can be seen in Fig. \ref{fig:lossvalue}. In this figure, the normalized loss values for uncertainty-based (ours) and uncertainty-less (mean-squared error) loss functions over each training sequence of the KITTI dataset are depicted on the Y-axis while the integration step is shown on the X-axis. Based on the mean of the normalized loss values over the dataset it can be seen that in the case of using a mean squared error as the loss function, the loss values increase exponentially as more terms are integrated. On the other hand, when using our approach, the weighting seen in Fig. \ref{fig:translationviolin} and \ref{fig:rotationviolin} does not allow the loss to increase exponentially and the increase in the loss magnitude exhibits a less aggressive behavior. A case study over a 10-step window is also provided in Fig. \ref{fig:casestudy}. It can be seen that frame to frame losses for both uncertainty-based and uncertainty-less losses for this short trajectory are highly correlated in terms of their behavior and the loss magnitude on the first iteration of the algorithm is the largest over the window. However, while the MSE loss increases exponentially with the introduction of integration, the uncertainty based loss does not exhibit the same behavior and rather than exponentially increasing, the precision term in the compounded loss (which is the result of the propagation of uncertainty) causes a decrease in the global loss term due to the large amount of uncertainty in the first step of the algorithm. This shows that the balanced weighting for the global and incremental loss terms in our approach requires no manual tuning or dataset-specific changes.

\begin{table}[!t]
    \centering
    \caption{Loop Closure Quantitative Analyses}
    \setlength{\tabcolsep}{0.55\tabcolsep}
    \begin{tabular}{c|cc|cc|cc} \toprule
        \multirow{3}{*}{Seq.} & \multicolumn{2}{c}{Baseline} & \multicolumn{2}{c}{VO} & \multicolumn{2}{c}{UVO (ours)} \\
               & t(\%) & r($^\circ$) & t(\%) & r($^\circ$) & t(\%) & r($^\circ$) \\
               & \multicolumn{2}{c}{(Aligned Traj.)} & \multicolumn{2}{c}{(Aligned Traj.)} & \multicolumn{2}{c}{(Aligned Traj.)}\\\midrule
        \multirow{2}{*}{13} & 5.109 & 2.210 & 8.084 & 4.125 & \textbf{3.395} & \textbf{1.390}\\
           & (5.725) & (2.210) & (8.116) & (4.125) & (\textbf{3.416}) & (\textbf{1.390})\\\midrule
        \multirow{2}{*}{15} & 14.20 & 3.465 & 10.49 & 1.706 & \textbf{10.18} & \textbf{1.330}\\
           & (9.813) & (3.465) & (5.135) & (1.706) & (\textbf{4.300}) & (\textbf{1.330})\\\midrule
        \multirow{2}{*}{Avg.} & 9.654 & 2.837 & 9.287 & 2.915 & \textbf{6.787} & \textbf{1.360} \\
           & (7.769) & (2.837) & (6.625) & (2.915) & (\textbf{3.858}) & (\textbf{1.360})\\
        \bottomrule
    \end{tabular}
    \label{tab:loopclosure}
\end{table}
\subsection{UVO and Loop Closure}
In this section, we use the incremental pose and uncertainty outputs of the network as the edges of a pose-graph to showcase the benefits of uncertainty estimation in a realistic scenario.
Moreover, we use DBoW3 \cite{galvez2012bags}, a loop detection algorithm based on bag-of-words representations of images, to define a similarity measure for pairs of images. When a loop is detected, an edge connects the corresponding nodes of images in the graph that are in the neighborhood of each other. Then, the pose and uncertainty of this edge are derived by passing this pair of frames to the network itself.
By solving this graph in different scenarios we may quantify the effectiveness of using uncertainty in such a setting. To form a baseline, we perform the same experiment once without any loops (termed baseline) and once with fixed uncertainty (termed VO) while the pose matrices are the network outputs. To perform this experiment we use sequences 13 and 15 of the KITTI dataset. Since the KITTI dataset does not provide a ground truth for these sequences, we used the stereo variant of ORB-SLAM2 \cite{mur2017orb}, which obtains an accuracy of 1.15\% on translation and 0.27$^\circ$ on rotation based on the KITTI odometry benchmark, as a reasonably accurate proxy for ground-truth.

The results from this experiment are provided in Table \ref{tab:loopclosure}. We report quantitative results in two scenarios. One where the output trajectories are untouched and one where the trajectories are scaled using Umeyama alignment \cite{umeyama1991least}. The latter scenario will allow us to evaluate the necessity of uncertainty prediction without any disruptions from scale errors. Based on the results from sequence 15, it can be seen that with the addition of loop closure, both uncertainty-based and uncertainty-less approaches provide a significant increase of 28.3\% and 26.1\% in translation accuracy over the untouched trajectories respectively. Meanwhile, the scaled trajectories show that the increase in the accuracy of uncertainty based estimates is 8.51\% larger compared to that of uncertainty-less study. On the other hand, based on the results from sequence 13, not using the estimated uncertainty values \emph{degrades} the accuracy of the algorithm by 58.2\% on translation and 86.6\% on rotation while using the estimated uncertainty allows for an increase in accuracy by 33.5\% on translation and 37.1\% on rotation. This is because the uncertainty-less experiment incorporates overconfident factors into the graph, while the uncertainty-based method balances the weights of the added factors. Overall, the mean accuracy of the loop closure enabled algorithm is increased by 29.7\% on translation and 52\% on rotation when using uncertainty values compared to the uncertainty-less study.

\begin{figure}%
\centering
\subfigure[Sequence 13]{%
\label{fig:lcseq13}%
\includegraphics[height=1.59in]{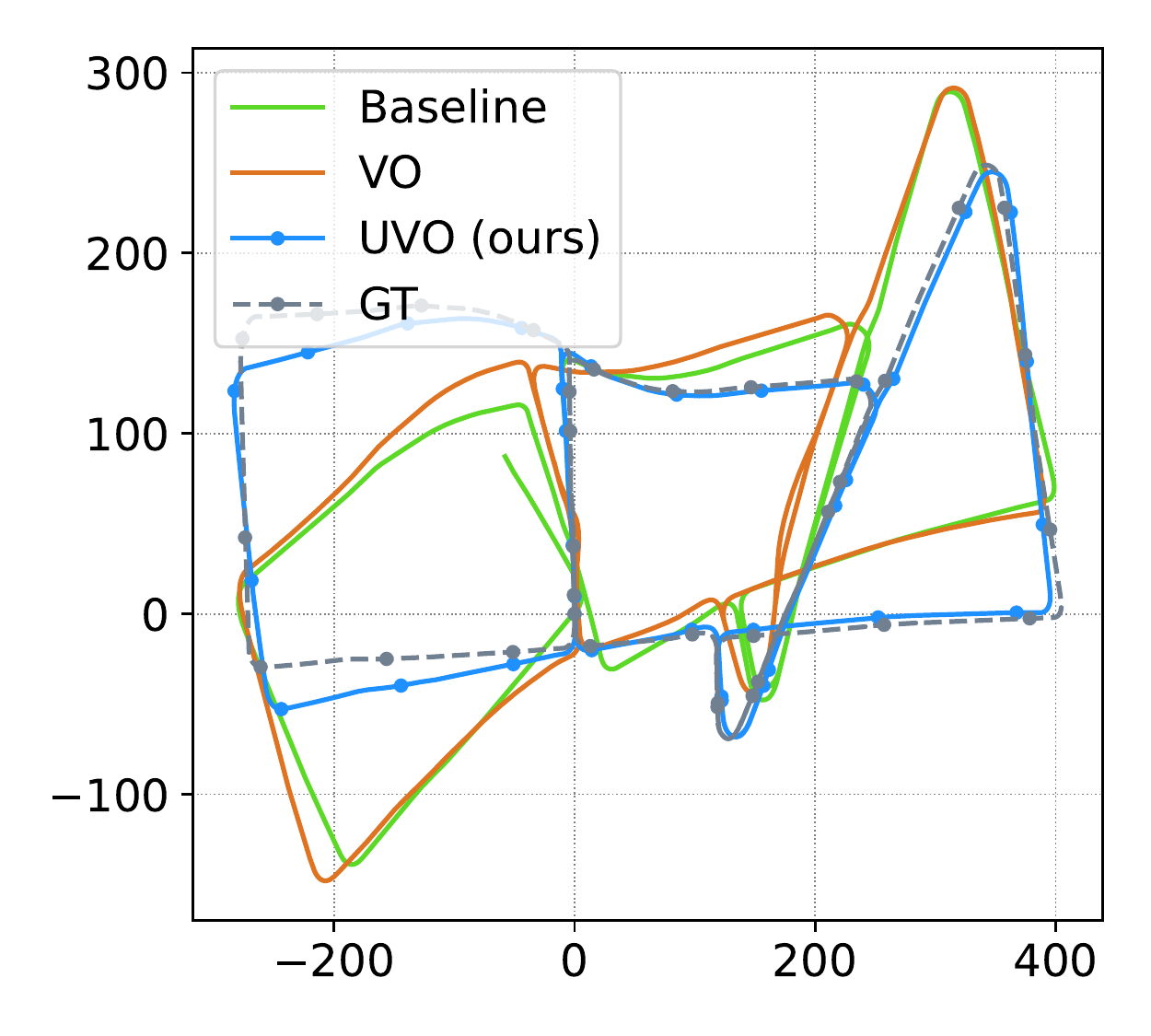}}%
\qquad
\hspace{-1.15cm}
\subfigure[Sequence 15]{%
\label{fig:lcseq15}%
\includegraphics[height=1.59in]{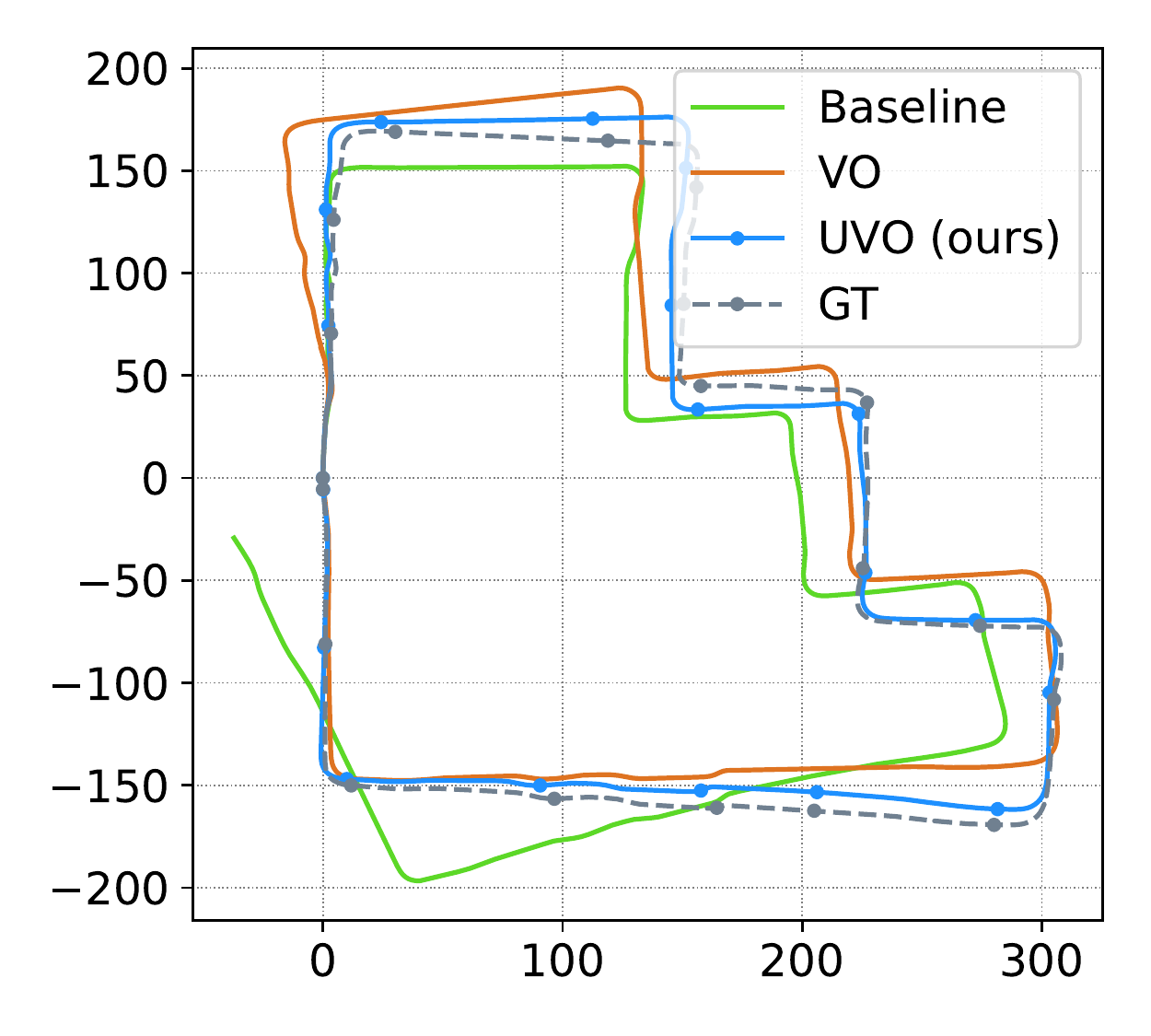}}%
\caption{Loop closure results on sequences 13 and 15 of the KITTI dataset (scale corrected)}
\label{fig:lcqualitative}
\vspace{-0.3cm}
\end{figure}

The resulting trajectories from this experiment are visualized in Fig. \ref{fig:lcqualitative}. In the case of sequence 15, it can be seen that at the start of the path (position (0, 50)) the outputs experience a large deviation from the ground-truth while the UVO outputs are able to track the true trajectory accurately. The results on sequence 13 are depicted in Fig. \ref{fig:lcseq15} and show that the estimated trajectory is able to closely follow the ground-truth trajectory especially in areas where loops are detected (the area that falls under $x>0$ in Fig. \ref{fig:lcseq15}) while uncertainty-less loop closure causes a degradation in the estimated trajectory.

\section{Conclusion}
This paper introduces a consistency-based loss function for deep odometry by compounding the estimated SE(3) pose and uncertainties. The compounded terms are then used in a negative log-likelihood objective function where the precision matrices weighting the global loss term are based on the integrated uncertainty. Quantitative and qualitative results against the SOTA in a visual odometry setting show that the addition of the proposed loss component allows our approach to significantly outperform the recently proposed SOTA methods in VO. Next, the estimated uncertainty values are evaluated and the mean uncertainty interval and out-of-range percentages are quantified to show that the output distribution adequately covers the ground-truth values. Then, the weighting resulted from the estimated precision matrices is visualized and the loss values from UVO are compared to the commonly used mean-squared error loss to show the appropriate balancing of the loss in case of our approach. Finally, the effectiveness of the estimated uncertainties is shown in a loop closure scenario where the constraints between the nodes are the pose and uncertainty estimates from our method. This analysis showed that the uncertainty estimates allow for a significant increase in accuracy while not using the estimated uncertainty to formulate the factors in the graph leads to a diminished accuracy.

\addtolength{\textheight}{-12cm}   





\bibliographystyle{IEEEtran}
\bibliography{root}

\end{document}